\documentclass[twoside]{article} \usepackage{aistats2017}

\usepackage[utf8]{inputenc} % allow utf-8 input
\usepackage[T1]{fontenc}    % use 8-bit T1 fonts
\usepackage{hyperref}       % hyperlinks
\usepackage{url}            % simple URL typesetting
\usepackage{booktabs}       % professional-quality tables
\usepackage{amsfonts}       % blackboard math symbols
\usepackage{nicefrac}       % compact symbols for 1/2, etc.
\usepackage{microtype}      % microtypography
\usepackage{algorithm} %ctan.org\pkg\algorithms
\usepackage{algpseudocode}
\usepackage{mathtools}

\usepackage{enumitem}
\usepackage{caption}
\usepackage{subcaption}
\usepackage[usenames]{color}
\newtheorem{theorem}{Theorem}
\newtheorem{definition}{Definition}

\newtheorem{proof}{Proof}

\providecommand{\nor}[1]{\left\lVert {#1} \right\rVert}

\begin{document}



\twocolumn[

\aistatstitle{Co-Occuring Directions Sketching for Approximate Matrix Multiply }
%\aistatsauthor{}
%\aistatsaddress{ }
\aistatsauthor{ Youssef Mroueh \And Etienne Marcheret  \And Vaibhava Goel }

\aistatsaddress{ IBM T.J Watson Research Center } ]

\begin{abstract}
We introduce  \emph{co-occurring directions} sketching, a deterministic  algorithm for approximate matrix product (AMM), in the streaming model.  We show that co-occuring directions achieves a better error bound for AMM than other randomized and deterministic approaches for AMM. Co-occurring directions gives a $(1+\varepsilon)$-approximation of the optimal low rank approximation of a matrix product. Empirically our algorithm outperforms competing methods for AMM, for a small sketch size. We validate empirically our theoretical findings and algorithms. %on both synthetic and real world multimodal datasets.  
\end{abstract}

\section{Introduction}
The vast and continuously growing amount of multimodal content poses some challenges with respect to the collection and the mining of this data.
Multimodal datasets are often viewed as multiple large matrices describing the same content with different modality representations (multiple views) such as images and their textual descriptions. The product of large  multimodal matrices is of practical interest as it models the correlation between different modalities. Methods such as Partial Least Squares (PLS) \cite{Wegelin00asurvey}, Canonical Correlation Analysis (CCA)\cite{HOTELLING}, Spectral Co-Clustering \cite{Dhillon}, exploit the low rank structure of the correlation matrix to mine the hidden joint factors, by computing the truncated singular value decomposition of a matrix product. 

The data streaming paradigm assumes a single pass over the data and a small memory footprint, resulting in a space/accuracy  tradeoff. Multimodal data can occupy a large amount of memory or may be generated sequentially, hence it is important  for the streaming model to capture the data correlation . 

Approximate Matrix Multiplication (AMM), is gaining an increasing interest in streaming applications (See the recent monograph \cite{Woodruff2014} for more details ). In AMM we are given matrices $X$,$Y$, with  a large number of columns $n$, and the goal is to compute matrices $B_{X},B_Y$, with smaller number of columns $\ell$, such that $||XY^{\top}-B_{X}B^{\top}_{Y}||_{Z}$ is small for some norm $\nor{.}_{Z}$. In streaming AMM, columns of $B_{X},B_{Y}$, need to be updated as the data arrives sequentially. We refer to $B_{X}$ and $B_{Y}$ as sketches of $X$ and $Y$.

Randomized approaches for AMM  were pioneered by the work of \cite{Drineas2006}. The approach of \cite{Drineas2006} is based on the \emph{sampling}  of $\ell$ columns of $X$ and $Y$.  \cite{Drineas2006} shows that by choosing an appropriate sampling matrix $\Pi \in \mathbb{R}^{n\times \ell}$, we obtain a Frobenius error guarantee ($\nor{.}_{Z}= \nor{.}_{F}$): 
\begin{equation}
\nor{XY^{\top}- X\Pi(Y \Pi)^{\top} }_{F}\leq \varepsilon \nor{X}_{F}\nor{Y}_{F},
\label{eq:sampling}
\end{equation}
for $\ell= \Omega(1/\varepsilon^2)$, with high probability.
The same guarantee of Eq. \eqref{eq:sampling} was achieved in \cite{Sarlos2006}, by using a \emph{random projection} $\Pi \in \mathbb{R}^{n\times \ell}$ that satisfies the guarantees of a Johnson- Lindenstrauss (JL) transform ($\forall x \in \mathbb{R}^n \nor{\Pi x}^2 \sim (1\pm \varepsilon)\nor{x}^2, \text{ with probability } 1-\delta$), where $\ell= O(1/\varepsilon^2\log(1/\delta))$.
Other randomized approaches focused on error guarantees given in spectral norm $(\nor{.}_{Z}=\nor{.})$ , such as JL embeddings  or  efficient subspace embeddings \cite{Sarlos2006,MZ11,AVZ14,CohenNW15} that can be applied to any type of matrices X in input sparisty time \cite{Clarkson2013}. \cite{CohenNW15} showed that  using a subspace embedding $\Pi \in \mathbb{R}^{n\times \ell}$ we have with a probability $1-\delta$:
\begin{equation}
\nor{XY^{\top}- X\Pi(Y \Pi)^{\top} }\leq \varepsilon \nor{X}\nor{Y},
\label{eq:subspace}
\end{equation}
for $\ell= O( (sr(X)+sr(Y)+\log(1/\delta) )/\varepsilon^2)$, where $sr(X)=\frac{\nor{X}^2_{F}}{\nor{X}^2}$ is the stable rank of $X$. Note that $sr(X)\leq rank(X)$, hence results stated in term of stable rank are sharper and more robust than the one stated with the rank \cite{Sarlos2006,MZ11,AVZ14}. 

Covariance sketching refers to AMM for $X=Y$. 
An elegant deterministic approach for covariance sketching called \emph{frequent directions} was introduced recently in \cite{Liberty,FD}, drawing the connection between \emph{covariance matrix sketching}, and the classic problem of estimation of \emph{frequent items} \cite{FT}.
Another approach for AMM, consists of concatenating matrices X and Y, and of applying a covariance sketch technique on the resulting matrix, this approach results in a looser guarantee; The right hand side in Equations \eqref{eq:sampling},\eqref{eq:subspace} is replaced by $\varepsilon (\nor{X}^2_{F}+\nor{Y}^2_{F})$. Based on this observation, \cite{YeLZ16} proposed to use the  \emph{frequent directions} algorithm of \cite {Liberty} to perform AMM in a deterministic way, we refer to this approach as FD-AMM. FD-AMM \cite{YeLZ16} outputs $B_{X},B_{Y}$ such that
\begin{equation}
\nor{XY^{\top }-B_{X}B^{\top}_{Y}}\leq \varepsilon (\nor{X}^2_{F}+\nor{Y}^2_{F}),
\label{eq:FDAMM}
\end{equation}
for $\ell =\lceil \frac{1}{\varepsilon}\rceil$. The sketch length $\ell$ dependency   on $\varepsilon$ in randomized methods is quadratic, FD-AMM improves this dependency to linear.

In this paper we introduce \emph{co-occuring directions}, a deterministic algorithm  for AMM. Our algorithm is  inspired by \emph{frequent directions}  and enables similar guarantees to \eqref{eq:subspace} in spectral norm, but with a linear dependency of $\ell$ on $\varepsilon$ as in FD-AMM. Given with stable ranks, co-occuring direction achieves the  guarantee of $\eqref{eq:subspace}$ for $\ell= O(\sqrt{sr(X)sr(Y)}/\varepsilon)$.

The paper is organized as follows: In Section \ref{sec:corrSketch} we review \emph{frequent directions}, introduce our \emph{co-occuring directions} sketching algorithm, and give error bounds analysis in AMM and in low rank approximation of a matrix product. We state our proofs in Section \ref{sec:proofs}. In section \ref{sec:discussion}  and Section \ref{sec:prev} we discuss error bounds, space and time requirements, and compare our approach to related work on AMM and low rank approximation. Finally we validate the empirical performance of  \emph{co-occuring directions}  in Section \ref{sec:experiments}, on both synthetic and real world multimodal datasets.

\paragraph{Notation.}
 We note by $C=U\Sigma V^{\top}$, the thin svd of $C$, and by $\sigma_{\max}(C)$ the maximum singular value, $Tr$ refers to the trace.
 $\sigma_j$ are the singular values that are assumed to be given in decreasing order.
Note that for $C \in \mathbb{R}^{m_x\times m_y}$ the spectral norm is  defined as follows 
 $\nor{C}= \max_{u,v,\nor{u}=\nor{v}=1}\left|u^{\top}C v\right|=\sigma_{\max}(C).$ % Note that we have  for any two matrices $B$ and $C$,
% $\nor{BC}\leq \nor{B}\nor{C},$ and for for any $x$, $\nor{Cx}_{2}\leq \nor{C}\nor{x}_{2}$.
The nuclear norm (known also as trace or $1-$ schatten norm) is defined as follows:
$\nor{C}_{*}= Tr(\Sigma).$ $sr(C)=\frac{\nor{C}^2_{F}}{\nor{C}^2}$ is the stable rank of $C$. Assume $C$ and $D$ have the same number of column, $[ C;D]$ denotes their concatenation on their row dimensions. 
For $n\in \mathbb{N}, [n]=\{1,\dots n\}$.
\section{Sketching from Covariance to Correlation} \label{sec:corrSketch}
In this section we review covariance sketching with the \emph{frequent directions} algorithm of \cite{Liberty} and state its  theoretical guarantees \cite{Liberty,FD}. We then introduce correlation sketching and present and analyze our \emph{co-occuring directions} algorithm.  
\subsection{Covariance Sketching: \emph{Frequent Directions}}
Let  $X \in \mathbb{R}^{m_x\times n}$, where $n$ is the number of samples and $m_x$ the dimension. We assume that $n>m_x$. The goal of covariance sketching is to find a small matrix $D_{X} \in \mathbb{R}^{m_x \times \ell}$, where $\ell<<n$ ($\ell$ is assumed to be an even number ), such that $XX^{\top}\approx D_{X}D_{X}^{\top}$. Frequent directions algorithm introduced in \cite{Liberty} (Algorithm \ref{ALG:fd}) achieves this  goal. Intuitively \emph{frequent directions} algorithm sets a noise level using the median of the spectrum of the covariance of the sketch $D_{X}$. It then discards directions below that level and  replaces them with fresh samples. This  results in the updated the covariance estimate. This process is repeated as the data is streaming.
\begin{algorithm}[H]
 \begin{algorithmic}[1]
 \Procedure{FD}{$X \in \mathbb{R}^{m_x \times n}$}
 \State  $D_{X} \gets 0 \in \mathbb{R}^{m_x \times \ell}$ .
 \For {$i \in [n]$}
 \State Insert column $X_i$ into a zero column of $D_{X}$
 \If {$D_{X}$ has no zero valued column} 
 \State $[U,\Sigma,V]\gets \text{SVD}(D_{X})$
 %\State $C \gets U \Sigma  $\Comment{not computed needed for proof notations}
 \State $\delta \gets \sigma^2_{\ell/2}$ \Comment{median value of $\Sigma^2$}
 \State $\tilde{\Sigma}\gets   \sqrt{\max(\Sigma^2-\delta I_{\ell},0)}$ \Comment{ shrinkage}
 \State $D_{X} \gets U\tilde{\Sigma}$
  \EndIf
 \EndFor
 \State \textbf{return} $D_{X}$
 \EndProcedure
 \end{algorithmic}
 \caption{Frequent Directions}
 \label{ALG:fd}
\end{algorithm}

\begin{theorem}[\cite{Liberty}] \label{theo:fd}

$D_{X}$ the output of algorithm \ref{ALG:fd} satisfies:
\begin{equation}
\nor{XX^{\top}-D_{X}D_{X}^{\top}} \leq \frac{2 \nor{X}^2_{F}}{ \ell}.
\label{eq:fdtheo}
\end{equation}
%\end{itemize}
\end{theorem}

\subsection{Correlation Sketching: \emph{Co-occuring Directions}}
We start by defining correlation sketching:
\begin{definition}[Correlation Sketching/AMM] Let $X \in \mathbb{R}^{m_x\times n}$, $Y \in \mathbb{R}^{m_y \times n}$, where $n>\max(m_x,m_y)$.
Let $B_{X} \in \mathbb{R}^{m_x \times \ell}$ and $B_Y \in \mathbb{R}^{m_y \times \ell}$ $(\ell <n, \ell \leq \min(m_x,m_y))$. Let $\eta>0$ . The matrix pair $(B_X,B_Y)$ is called an $\eta$-correlation sketch of $(X,Y)$ if it satisfies in spectral norm:
$$\nor{XY^{\top}-B_{X}B_{Y}^{\top}}\leq \eta.$$
\label{def:corS}
\end{definition}
\vskip -0.3 in
We now  present our \emph{co-occuring directions} algorithm (Algorithm \ref{ALG:CorrSketch}). Intuitively Algorithm \ref{ALG:CorrSketch} sets a noise level using the median of the singular values of the correlation matrix of the sketch $B_{X}B_{Y}^{\top}$. The SVD of $B_{X}B_{Y}^{\top}$ is computed efficiently in lines 8,9 and 10 of Algorithm \ref{ALG:CorrSketch} using QR decomposition. Left and right singular vectors below this noise threshold are replaced by fresh samples from $X$ and $Y$, correlation sketches are updated and the process continues.
Theorem \ref{theo:SketchCorr} shows that our \emph{co-occuring directions} algorithm outputs $(B_X,B_Y)$ a correlation sketch of $(X,Y)$ as defined above in Definition \ref{def:corS}. 

\begin{algorithm}[H]
 \begin{algorithmic}[1]
 \Procedure{Co-D}{$X \in \mathbb{R}^{m_x \times n},Y \in \mathbb{R}^{m_y \times n}$}
 \State  $B_{X} \gets 0 \in \mathbb{R}^{  m_x \times \ell}$ .
  \State  $B_{Y} \gets 0 \in \mathbb{R}^{m_y \times \ell}$ .

 \For {$i \in [n]$}
 \State Insert a column $X_i$ into a zero valued column  of $B_X$
  \State Insert a column $Y_i$ into a zero valued column of $B_Y$
 \If {$B_X,B_Y$ have no zero valued column} 
 \State $[Q_x,R_x]\gets \text{QR}(B_{X})$
  \State $[Q_y,R_y]\gets \text{QR}(B_{Y})$ 
\State $[U,\Sigma,V]\gets \text{SVD}( R_xR_y^{\top})$ \\ 
\Comment{$Q_x \in \mathbb{R}^{m_x\times \ell},R_x\in \mathbb{R}^{\ell \times \ell}$,} \\
\Comment{$Q_y \in \mathbb{R}^{m_y\times \ell},R_y\in \mathbb{R}^{\ell \times \ell}$,
$U,\Sigma,V \in \mathbb{R}^{\ell \times \ell}$. }
%\State $C= V_1U \Sigma V^{\top}V_2$\Comment{not computed needed for proof notations}
\State $C_x \gets  Q_xU\sqrt{{\Sigma}}$
\State $C_y \gets  Q_yV\sqrt{{\Sigma}}$\\ \Comment{ $C_x,C_y$ not computed }
 \State $\delta \gets \sigma_{\ell/2}(\Sigma)$ \Comment{the median value of $\Sigma$}
 \State $\tilde{\Sigma}\gets  {\max(\Sigma-\delta I_{\ell},0)} $ \Comment {shrinkage}
% \State $\tilde{C} =V_1 U \tilde{\Sigma} V^{\top}V_2 $ \Comment{not computed needed for proof notations}
 \State $B_X \gets Q_xU \sqrt{\tilde{\Sigma}}$
 \State  $B_Y \gets Q_yV\sqrt{\tilde{\Sigma}}$\\
 \Comment{at least last $\ell/2$ columns are zero}
  \EndIf
 \EndFor
 \State \textbf{return} $B_X,B_Y$
 \EndProcedure
 \end{algorithmic}
 \caption{Co-occuring Directions }
 \label{ALG:CorrSketch}
\end{algorithm}

It is important to see that while frequent directions shrinks $\Sigma^2$, co-occuring directions filters $\Sigma$.
We prove in the following an approximation bound in spectral norm for co-occurring directions.
\subsubsection{Main Results}
We give in the following our main results, on the approximation error of co-occurring direction in AMM (Theorem \ref{theo:SketchCorr}), and in the $k-$th rank approximation of a matrix product (Theorem \ref{theo:LR}). Proofs are given in Section \ref{sec:proofs}.

\begin{theorem}[AMM]
The output of \emph{co-occuring} directions (Algorithm \ref{ALG:CorrSketch}) gives a correlation sketch $(B_X,B_Y)$ of $(X,Y)$, for $\ell \leq \min(m_x,m_y)$ satisfying:

For a correlation sketch of length $\ell$, we have:
 $$\nor{XY^{\top}- B_X B^{\top}_Y}\leq \frac{2 \nor{X}_{F}\nor{Y}_{F} }{ \ell}. $$

2) Algorithm \ref{ALG:CorrSketch} runs in $O(n(m_x+m_y+\ell)\ell)$ time and requires a space of  $O((m_x+m_y+\ell)\ell)$.\\

\label{theo:SketchCorr}
\end{theorem}

\begin{theorem}[Low Rank  Product Approximation] Let $(B_X,B_Y)$ be the output of  Algorithm \ref{ALG:CorrSketch}. Let $k \leq \ell$. Let $U_k,V_k$ be the matrices whose columns are the k-th largest left and right singular vectors of $B_{X}B_{Y}^{\top}$. 
Let $\pi^{k}_{U}(X)=U_kU_k^{\top}X, \pi^{k}_{V}(Y)=V_kV_k^{\top}Y$. Let $\varepsilon >0$, for $\ell \geq 8\frac{ \sqrt{sr(X)sr(Y)}}{ \varepsilon} \frac{||X|| ||Y||}{\sigma_{k+1}(XY^{\top})}$ we have:
$\nor{XY^{\top}-\pi^{k}_{U}(X)\pi^{k}_{V}(Y)^{\top}} \leq  \sigma_{k+1}(XY^{\top})(1+\varepsilon )$.
\label{theo:LR}
\end{theorem}
\subsubsection{Discussion of Main Results}\label{sec:discussion}
 For $\ell =\lceil \frac{1}{\varepsilon}\rceil, \varepsilon \in [\frac{1}{\min(m_x,m_y)},1]$ from Theorem \ref{theo:SketchCorr} we see that $(B_X,B_Y)$ produced by Algorithm \ref{ALG:CorrSketch} is an $\eta$-correlation sketch of $(X,Y)$ for $\eta=2 \varepsilon \nor{X}_{F}\nor{Y}_{F}$. In AMM, bounds are usually stated  in term of the product of spectral norms of $X$ an $Y$ as in Equation \eqref{eq:subspace}. Let $sr(X)=\frac{\nor{X}^2_{F}}{\nor{X}^2}$ be the stable rank of $X$. It is easy to see that co-occuring directions for $\ell =\frac{2\sqrt{sr(X)sr(Y)}}{\varepsilon  }$, gives an error bound of $\varepsilon \nor{X}\nor{Y}$. While in randomized methods the  error is $O(1/\sqrt{\ell})$, co-occuring direction's error is $O(1/\ell)$. Moreover the dependency on stable ranks in co-occuring directions is $2\sqrt{sr(X)sr(Y)}\leq sr(X)+sr(Y)$, the lattter appears in subspace embedding based AMM \cite{CohenNW15,MZ11,AVZ14}. 
 For $X=Y$ \emph{co-occuring directions} reduces to \emph{frequent directions} of \cite{Liberty}, and Theorem \ref{theo:SketchCorr} recovers Theorem \ref{theo:fd} of \cite{Liberty}.
 
 Stronger bounds for frequent directions were given in \cite{FD} where the bound in Equation \eqref{eq:fdtheo} is improved, for $\ell>2k$, for any $k$:
$${\nor{XX^{\top}-D_{X}D_{X}^{\top}} \leq \frac{2}{ \ell-2k} \nor{X-X_k}^2_{F},}$$
 where $X_{k}$ is the $k-$th rank approximation of $X$ (with $X_0=0$). Hence by defining $Z=[X;Y]\in \mathbb{R}^{(m_x+m_y)\times n}$ and applying frequent directions to $Z$ (FD-AMM \cite{YeLZ16}), we obtain $B_{X},B_{Y}$  satisfying:
  ${\nor{XY^{\top}-B_{X}B_{Y}^{\top}} \leq \frac{2}{\ell-2k} \nor{Z-Z_k}^2_{F},}$ hence the perfomance of FD-AMM depends on the low rank structure of $Z$. A sharper analysis for co-occuring directions remains an open question, but the following discussion of Theorem \ref{theo:LR} will shed some light on the advantages of co-occuring directions on FD-AMM \cite{YeLZ16}.

Theorem \ref{theo:LR} shows that co-occuring directions sketching gives  a $(1+\varepsilon)$- approximation of the optimal low rank approximation of the matrix product $XY^{\top}$. 
Note that $\sigma_{k+1}(XY^{\top}) \leq \frac{\nor{XY^{\top}}_*}{k+1}$. Hence  for $\ell\geq 8(k+1)/\varepsilon$, we obtain a $1+\varepsilon$- approximation of the optimal $k$ rank approximation of $XY^{\top}$. This highlights the relation between the sketch length in co-occurring directions $\ell$ and the rank  of $XY^{\top}$. Note that the maximum rank of $XY^{\top}$ is $\min(rank(X),rank(Y))$. When using FD-AMM, based on the covariance sketch of the concatenation of $X$ and $Y$, the sketch length $\ell$ is related to the rank of $Z=[X;Y]$. Note that the maximum rank of the concatenation ($Z$) is bounded by $rank(X)+rank(Y)$. Hence we see that co-occuring directions guarantees a $1+\varepsilon$ approximation of the optimal $k$-rank approximation of $XY^{\top}$ for a smaller sketch size then FD-AMM ($\min(rank(X),rank(Y))$ for \emph{co-occuring directions} versus $rank(X)+rank(Y)$ for FD-AMM).

In the following we comment on the running time of co-occuring directions.
\vskip -0.6 in
\subsubsection{Running Time Analysis and Parralelization.}\label{sec:runningtime}

\textbf{Running Time.} We compare the space and the  running time of our  sketch to to a naive implementation of the correlation sketch.\\
\emph{1) Naive Correlation Sketch}: In the if statement of Algorithm \ref{ALG:CorrSketch}, compute the $\ell$ thin svd  $\text{SVD}(B_{X}B^{\top}_Y)=[U,\Sigma,V]$, $B_{X}\gets  U\sqrt{\tilde{\Sigma}}, B_{Y}\gets V\sqrt{\tilde{\Sigma}} $. We need a space $O(m_xm_y)$ to store $B_{X} B_y^{\top}$. The running time is dominated by computing an $\ell$ thin svd $O(m_xm_y\ell)$ each $\frac{n}{\ell/2}$ that is $O(nm_xm_y )$, hence no gain with respect to brute force.\\
\emph{2) Co-occuring Directions}: Algorithm \ref{ALG:CorrSketch} avoids computing $B_{X}B_{Y}^{\top}$ by using the QR decomposition of $B_{X}$ and $B_{Y}$. The space needed is $O(\ell(m_x+m_y+\ell))$. We have a computation done every $\frac{n}{\ell/2}$, that is dominated by  computing QR factorization and svd : $O((m_x+m_y+\ell)\ell^2)$ (computing $R_xR_y^{\top}$ requires $O(\ell^3)$ operations). This results in  a total running time : 
$O(n(m_x+m_y+\ell)\ell)$. There is a computational and memory advantage when $\ell< \frac{m_xm_y}{m_x+m_y}$.

\textbf{Parallelization of Co-occuring Directions (Sketches of Sketches).} Similarly to the frequent directions \cite{Liberty}, co-occuring directions algorithm is simply parallelizable. Let $X=[X_1,X_2]\in \mathbb{R}^{m_x\times (n_1+n_2)}$, and $Y=[Y_1,Y_2] \in \mathbb{R}^{m_x\times (n_1+n_2)}$. Let $(B^1_X,B^1_Y)$ be the correlation sketch of $(X_1,Y_1)$, and 
$(B^2_X,B^2_Y)$ be the correlation sketch of $(X_2,Y_2)$. Then the correlation sketch $(C_X,C_Y)$ of $([B^1_X,B^2_X],[B^1_Y,B^2_Y])$ is a correlation sketch of $(X,Y)$, and is as good as $(B_X,B_Y)$ the correlation sketch of $(X,Y)$. Hence we can sketch the data in $M$-independent chunks on $M$ machines then merge by concatenating the sketches and performing another sketch on the concatenation, by doing so we divide the running time by $M$.

\section{Proofs}\label{sec:proofs}
In this Section we give proofs of our main results:

\begin{proof}[Proof of Theorem \ref{theo:SketchCorr}]
\noindent By construction we have:
\begin{eqnarray*}
C_xC^{\top}_y&=&\left(Q_xU\sqrt{{\Sigma}}\right) \left(Q_yV\sqrt{{\Sigma}}\right)^{\top} \\
&=& Q_x \left(U\Sigma V^{\top}\right)Q_{y}^{\top}
= Q_x\left(R_xR_y^{\top}\right)Q_y^{\top}\\
&=& \left(Q_xR_x\right)\left(Q_yR_y\right)^{\top}.
\end{eqnarray*}
Hence the algorithm is computing a form of  R-SVD of $B_X B^{\top}_Y$, followed by a shrinkage of the correlation matrix.
Let $B^{i}_x, B^{i}_y , C^{i}_x,C^{i}_y , \Sigma^i, \tilde{\Sigma^i},\delta_i$, the values of $ B_X, B_Y , C_x,C_y , \Sigma, \tilde{\Sigma},\delta$ after the execution of the main loop. $\delta_i =0$ if we don't enter the if statement ($B^i_x=C^i_x$ and $B^i_y=C^i_y$ if we don't enter the if statement).\\
Hence we have at an iteration $i$:
$$C^{i}_xC^{i,\top}_y= B^{i-1}_xB^{i-1,\top}_y+ X_{i} Y_{i}^{\top}.$$
Note that:
\begin{eqnarray*}
 XY^{\top} - B_XB^{\top}_Y &=&   XY^{\top} - B^{n}_{x}B^{n,\top}_{y}  \\
&=&    \sum_{i=1}^n \left(X_{i} Y_{i}^{\top}+ B^{i-1}_{x}B^{i-1,\top}_{y}  - B^{i}_{x}B^{i,\top}_{y} \right) \\
&=& \sum_{i=1}^n \left( C^{i}_xC^{i,\top}_y - B^{i}_{x}B^{i,\top}_{y} \right).
 \end{eqnarray*}
By the triangular inequality we can bound the spectral norm: 
$$ \nor{  XY^{\top} - B_XB^{\top}_Y } \leq \sum_{i=1}^n \nor{ C^{i}_xC^{i,\top}_y - B^{i}_{x}B^{i,\top}_{y} }.$$
We are left with bounding $\nor{C^{i}_xC^{i,\top}_y - B^{i}_{x}B^{i,\top}_{y}}$: 
$$C^{i}_xC^{i,\top}_{y}=  \left(Q^i_x U^i\right) \Sigma^i \left(Q^i_yV^i\right)^{\top}, B^{i}_xB^{i,\top}_{y}=  \left(Q^i_x U^i\right) \tilde{\Sigma}^i \left(Q^i_yV^i\right)^{\top}.$$
Note that: %for singular values we have: $\sigma_{i}(AB)\leq \nor{A}\sigma_{i}(B)$.
 %Then we have:
\begin{eqnarray*}
\nor{C^{i}_xC^{i,\top}_y - B^{i}_{x}B^{i,\top}_{y}} &=& \nor{(Q^i_x  U^i) (\Sigma^i-\tilde{\Sigma}^i)(Q^i_yV^i)^{\top}}\\
&=& \nor{\Sigma^i-\tilde{\Sigma}^i}\\
&\leq& \delta_i,
\end{eqnarray*}
where the  first equality follows from the fact that,  $Q^i_x U^i, Q^i_yV^i,$ are orthonormal. And $\Sigma^i-\tilde{\Sigma}^i$ is a diagonal matrix with at least $\ell/2$ entries equal $\delta_i$ or $0$, and the other entries are less than $\delta_i$.
% \begin{eqnarray*}
% \left| u^{\top}C^{\top}_{1}C_{2} v - u^{\top}B^{\top}_{1}B_{2} v\right|&=& \left| u^{\top} \left(V_1\Sigma_1^{\top}U \Sigma V^{\top}\Sigma_2V^{\top}_2\right) v -u^{\top} \left(V_1\tilde{\Sigma}_1^{\top}U \tilde{\Sigma} V^{\top}\tilde{\Sigma_2}V^{\top}_2\right) v \right| \\
%&=& \left| u^{\top} \left(V_1(\Sigma_1^{\top} -\tilde{\Sigma}_1)U \Sigma V^{\top}\Sigma_2V^{\top}_2\right) v -u^{\top} \left(V_1\tilde{\Sigma}_1^{\top}U \tilde{\Sigma} V^{\top}\tilde{\Sigma_2}V^{\top}_2\right) v \right|\\
%&\leq&\nor{u^{\top}V_1\Sigma_1^{\top}U}_2 \nor{\Sigma-\tilde{\Sigma}} \nor{V^{\top}\Sigma_2V^{\top}_2 v}_2\\
%&=& \nor{\Sigma-\tilde{\Sigma}} \nor{u^{\top}V_1\Sigma_1^{\top}U^{\top}_1}_2\nor{U_2\Sigma_2V^{\top}_2 v}_2 \text{ Since $U,V,U_1,U_2$ are othonormal }\\
%&\leq& {\delta} \nor{B_1 u}_2  \nor{B_2 v}_2.
% \end{eqnarray*}
It follows that we have in spectral norm:
\begin{equation}
 \nor{  XY^{\top} - B_XB^{\top}_Y }  \leq \sum_{i=1}^n \delta_i.
\label{eq:BSpectral}
\end{equation}
\vskip -0.2 in 
Now we want to relate $\sum_{i=1}^n \delta_i$ to $\ell$, and propreties of $X,Y$.\\
Let $\nor{.}_{*}$, the $1-$ schatten norm. For a matrix $A$ of rank $r$, and singular values $\sigma_i$ :
$\nor{A}_{*}=\sum_{i=1}^r \sigma_i(A).$
%For $A, E \in \mathbb{R}^{m,n} $, we have: 
%$$\left|\sigma_{i}(A+E) -\sigma_{i}(A)\right| \leq \nor{E}$$
We have:
\begin{eqnarray}
\nor{B_XB^{\top}_Y}_{*}&=&\nor{B^{n}_xB^{n,\top}_y}_{*}\nonumber \\&=& \sum_{i=1}^n \nor{B^{i}_xB^{i,\top}_y}_{*} - \nor{B^{i-1}_xB^{i-1,\top}_y}_{*} \nonumber \\
&=& \sum_{i=1}^n \left( \nor{C^{i}_xC^{i,\top}_y}_{*}- \nor{B^{i-1}_xB^{i-1,\top}_y}_{*} \right)\nonumber\\ &-& \sum_{i=1}^n \left(\nor{C^{i}_xC^{i,\top}_y}_{*}- \nor{B^{i}_xB^{i,\top}_y}_{*} \right)
\label{eq:Bound}
\end{eqnarray} 
 We have at an iteration $i$, the R-SVD of $C^{i}_xC^{i,\top}_y$ and $B^{i,}_xB^{i,\top}_y$:
$$ \nor{C^{i}_xC^{i,\top}_y}_{*}=Tr(\Sigma^i) \text{ and } \nor{B^{i}_xB^{i,\top}_y}_{*}=Tr(\tilde{\Sigma}^i).$$
%$$B^{i,}_xB^{i,\top}_y= (Q^i_x  U^i) \tilde{\Sigma}^i (Q^{i}_yV^i)^{\top},$$
Hence we have by the definition of the shrinking operation:  $\nor{C^{i}_xC^{i,\top}_y}_{*}- \nor{B^{i}_xB^{i,\top}_y}_{*}$
\begin{eqnarray}
&=&Tr(\Sigma^i-\tilde{\Sigma}^i)= \sum_{j=1}^{\ell} \sigma^i_j -\tilde{\sigma}^i_j\nonumber\\
& =&\sum_{j, \sigma^i_j> \delta_i} \delta_i + \sum_{j, \sigma^i_{j}\leq \delta_i} \sigma^i_{j} \geq  \frac{\ell}{2} \delta_i.
\label{eq:Tresh}
\end{eqnarray}
On the other hand using the reverse triangle inequality for the $1-$ shatten norm we have:
$$\nor{C^{i}_xC^{i,\top}_y}_{*}-\nor{B^{i-1}_xB^{i-1,\top}_y}_{*} \leq \nor{C^{i}_xC^{i,\top}_y-B^{i-1}_x B^{i-1,\top}_y }_{*}$$
Recall that:
$C^{i}_xC^{i,\top}_y = B^{i-1}_xB^{i-1,\top}_y+ X_{i} Y_{i}^{\top},$
hence we have:
\begin{equation}
\nor{C^{i}_xC^{i,\top}_y}_{*}-\nor{B^{i-1}_xB^{i-1,\top}_y}_{*} \leq \nor{X_{i} Y_{i}^{\top}}_{*}= \nor{X_{i}}_{2}\nor{Y_{i}}_2 ,
\label{eq:ReverseTriangIneq}
\end{equation}
since $X_{i}Y_{i}^{\top}$ is rank one. 
Finally putting together Equations \eqref{eq:Bound}, \eqref{eq:Tresh},\eqref{eq:ReverseTriangIneq}, we have:
\begin{equation}
\nor{B_XB^{\top}_Y}_{*} \leq \sum_{i=1}^{n} \nor{X_{i}}_{2}\nor{Y_{i}}_2 -  \frac{\ell}{2}\sum_{i=1}^n \delta_i.
\label{eq:Bdelta}
\end{equation}
It follows from Equation \eqref{eq:Bdelta} that:
\begin{eqnarray}
\sum_{i=1}^n \delta_i &\leq& \frac{2}{ \ell}\left(\sum_{i=1}^{n} \nor{X_{i}}_{2}\nor{Y_{i}}_2 - \nor{B_XB^{\top}_Y}_{*} \right) \nonumber\\
&\leq&   \frac{2}{\ell} \left(\sqrt{\sum_{i=1}^{n} \nor{X_{i}}^2_{2}} \sqrt{\sum_{i=1}^n \nor{Y_{i}}^2_2 }\right) \nonumber \\
&=&   \frac{2}{ \ell} \nor{X}_{F}\nor{Y}_{F},
\label{eq:fdeltabound}
\end{eqnarray}
where in the last inequality we used the Cauchy-Schwarz inequality.
Putting together Equations \eqref{eq:BSpectral} and  \eqref{eq:fdeltabound}  we have finally:
\begin{equation}
 \nor{  XY^{\top} - B_XB^{\top}_Y }  \leq  \frac{2}{ \ell} \nor{X}_{F}\nor{Y}_{F}.
\end{equation}
2) Refer to Section \ref{sec:runningtime}.
\end{proof}
\vskip -0.3 in
\begin{proof}[Proof of Theorem \ref{theo:LR}]

%We have $\nor{X-U_kU_k^{\top}X}^2_{F}+  \nor{Y-V_kV_k^{\top}Y}^2_{F}$
%\begin{eqnarray*}
%&=& \nor{X}^2_{F} -\nor{U_kU_k^{\top}X}^2_{F} +\nor{Y}^2_{F}-\nor{V_kV_k^{\top}Y}^2_{F}\\
%&=&  \nor{X}^2_{F}+\nor{Y}^2_{F}-\sum_{i=1}^k \nor{u_i^{\top}X}^2-\sum_{i=1}^k \nor{v_i^{\top}Y}^2\\
%&=& \nor{X}^2_{F}+\nor{Y}^2_{F} - \sum_{i=1}^k \nor{u_i^{\top}X- v_i^{\top}Y}^2 \\&-2& \sum_{i=1}^k u^{\top}_iXY^{\top}v_{i}\\
%&\leq&  \nor{X}^2_{F}+\nor{Y}^2_{F} -2 \sum_{i=1}^k u^{\top}_iXY^{\top}v_{i}\\
%&=&  \nor{X}^2_{F}+\nor{Y}^2_{F} +2 \sum_{i=1}^k u^{\top}_i(B_{X}B_{Y}^{\top}-XY^{\top})v_{i} \\
%&-& \underbrace{2 \sum_{i=1}^{n}u^{\top}_iB_{X}B_{Y}^{\top}v_i}_{\geq 0}\\
%&\leq& \nor{X}^2_{F}+\nor{Y}^2_{F} + 2 k \nor{XY^{\top}-B_{X}B_{Y}^{\top}}\\
%&\leq& \nor{X}^2_{F}+\nor{Y}^2_{F}+ 2\nor{X}_{F}\nor{Y}_{F}\frac{2k}{\alpha \ell}.
%\end{eqnarray*}

Let $\pi^{k}_{U}(X)=U_kU_k^{\top}X, \pi^{k}_{V}(Y)=V_kV_k^{\top}Y$.
Let $\mathcal{H}^x_{k}$ be the span of $\{u_1,\dots u_{k}\}$, and $\mathcal{H}^x_{m_x-k}$ be the orthogonal of  $\mathcal{H}^x_{k}$.
Similarly define  $\mathcal{H}^y_{k}$ the span of $\{v_1,\dots v_{k}\}$, and $\mathcal{H}^y_{m_y-k}$ its orthogonal. 
For all $u \in \mathbb{R}^{m_x}, \nor{u}=1$, there exits $a_{x}, b_{x} \in \mathbb{R}$, $a^2_{x}+ b^2_{x} =1$, such that $u=a_{x} w_{x}+ b_{x} z_{x}$, 
where $w_x \in \mathcal{H}^x_{k},||w_x||=1$ and $z_{x} \in \mathcal{H}^x_{m_x-k}, ||z_{x}||=1 $. 
Similarly for $v \in  \mathbb{R}^{m_y}, \nor{v}=1$ there exits $a_{y}, b_{y} \in \mathbb{R}$, $a^2_{y}+ b^2_{y} =1$, such that $v=a_{y} w_{y}+ b_{y} z_{y}$, 
where $w_y \in \mathcal{H}^y_{k}, ||w_y||=1$ and $z_{y} \in \mathcal{H}^y_{m_y-k},||v_y||=1 $ . 

Let $\Delta = XY^{\top}-\pi^{k}_{U}(X)\pi^{k}_{V}(Y)^{\top}$, we have $\nor{\Delta} = \max_{u\in \mathbb{R}^{m_x},v\in \mathbb{R}^{m_y},||u||=||v||=1}|u^{\top}\Delta v|$
\begin{eqnarray*}
 |u^{\top}\Delta v| &=& |(a_{x} w_{x}+ b_{x} z_{x})^{\top}\Delta (a_{y} w_{y}+ b_{y} z_{y})|\\
 &\leq& |a_{x}a_{y}| |w_{x}^{\top}\Delta w_{y}| +|b_{x}b_{y}| |z_{x}^{\top}\Delta z_y | \\
 &+&  |a_{x}b_{y}| |w_x^{\top} \Delta z_{y} |+ |b_{x}a_{y}| |z_x^{\top} \Delta w_{y}|
\end{eqnarray*}
Since $w_x \in \mathcal{H}^x_{k},w_y \in \mathcal{H}^y_{k}, $ we have $w_{x}^{\top}\Delta w_{y}= 0$.
Since $z_x \in \mathcal{H}^x_{m_x-k},z_y \in \mathcal{H}^y_{m_y-k}, z_{x}^{\top}\Delta z_y = z_{x}^{\top}XY^{\top}z_y$. Similarly $w_x^{\top} \Delta z_{y}= w_x^{\top}XY^{\top}z_{y}$, and $ z_x^{\top} \Delta w_{y}= z_x^{\top} XY^{\top} w_{y}$. Note that $|a_x|, |b_x|,|a_y|,|b_y|$ are bounded by 1. Hence we have (maximum is taken on each appropriate set defined above, all vectors are unit norm):
\begin{eqnarray*}
 \max_{u,v } |u^{\top}\Delta v|&\leq& \max_{z_x,z_y} |z_{x}^{\top}XY^{\top}z_y|+ \max_{w_x,z_y}|w_x^{\top}XY^{\top}z_{y}|\\
 &+&\max_{z_x,w_y} |z_x^{\top} XY^{\top} w_{y}|
 \end{eqnarray*}
For  $z_x \in \mathcal{H}^x_{m_x-k},z_y \in \mathcal{H}^y_{m_y-k}$ we have:
 \begin{eqnarray*}
 |z_{x}^{\top}XY^{\top}z_y| &\leq&| z_{x}^{\top}(XY^{\top}-B_XB_Y^{\top})z_y| + |z_{x}^{\top} B_{X}B_{Y}^{\top}z_{y}|\\
 &\leq& \nor{XY^{\top}-B_XB_Y^{\top}}+\sigma_{k+1}(B_{X}B_{Y}^{\top})\\
 &\leq& 2\nor{XY^{\top}-B_XB_Y^{\top}} +\sigma_{k+1}(XY^{\top}),
 \end{eqnarray*}
where we used that $\max_{z_x \in \mathcal{H}^x_{m_x-k},z_y \in \mathcal{H}^y_{m_y-k}}|z_{x}^{\top} B_{X}B_{Y}^{\top}z_{y}|$\\$=\sigma_{k+1}(B_{X}B_{Y}^{\top})$ by definition of $\sigma_{k+1}$. The last inequality follows from weyl inequality $|\sigma_{k+1}(B_{X}B_{Y}^{\top}) -\sigma_{k+1}(XY^{\top})|\leq \nor{XY^{\top}-B_XB_Y^{\top}}$.

Note that for $w_x \in \mathcal{H}^x_{k}$ and $z_y \in \mathcal{H}^{y}_{m_y-k}$ we have $w_{x}^{\top}B_{X}B_{Y}^{\top}z_{y} =0$. 
To see that, note that $w_x \in span\{u_1,\dots u_k\}$ , $z_y \in span\{v_{k+1},\dots v_{\ell}\}$. There exists $\beta _j$, such that $z_{y}=\sum_{j=k+1}^{\ell}\beta_j v_j$, hence  $B_{X}B_{Y}^{\top}z_y=\sum_{i=1}^{\ell}\sum_{j=k+1}^{\ell}\sigma_{i}\beta_j u_iv_{i}^{\top}v_j=\sum_{j=k+1}^{\ell}\sigma_{j}\beta_{j}u_j \perp w_{x}$. 
Hence we have: 
\begin{eqnarray*}
|w_x^{\top}XY^{\top}z_{y}|&=&|w_x^{\top}(XY^{\top}-B_{X}B_{Y}^{\top})z_{y}|\\
&\leq& \nor{XY^{\top}-B_{X}B_{Y}^{\top}}.
\end{eqnarray*}
\vskip -0.2 in 
Similarly for  for $z_x \in \mathcal{H}^x_{m_x-k}$ and $w_y \in \mathcal{H}^{y}_{k}$ we conclude that:
$|w_x^{\top}XY^{\top}z_{y}|\leq  \nor{XY^{\top}-B_{X}B_{Y}^{\top}} $.
Finally we have:
\begin{eqnarray*}
\nor{\Delta }&\leq& 4 \nor{XY^{\top}-B_{X}B_{Y}^{\top}} + \sigma_{k+1}(XY^{\top})\\
&\leq&\frac{8\nor{X}_{F}\nor{Y}_{F} }{\ell}+ \sigma_{k+1}(XY^{\top})\\
&\leq& \sigma_{k+1}(XY^{\top}) (1+ 8\frac{ \sqrt{sr(X)sr(Y)}}{ \ell} \frac{||X|| ||Y||}{\sigma_{k+1}(XY^{\top})})
\end{eqnarray*}
For $\ell \geq 8\frac{ \sqrt{sr(X)sr(Y)}}{ \varepsilon} \frac{||X|| ||Y||}{\sigma_{k+1}(XY^{\top})}$, we have:
$\nor{\Delta } \leq \sigma_{k+1}(XY^{\top})(1+\varepsilon)  $.

\end{proof}

\section{Previous Work on Approximate Matrix Multilply}\label{sec:prev}
We list here a catalog of baselines  for AMM: 

 \textbf{Brute Force.} We keep a running correlation $C \gets C+ X_i Y^{\top}_i$. We perform an $\ell$ thin svd at the end of the stream. Space $O(m_x m_y)$, 
running time: ${O(n m_x m_y)}+O(m_x m_y \ell),$ the cost of the sketch update and the $\ell$ thin svd.

\textbf{Sampling }\cite{Drineas2006}. We define a distribution over $[n]$, $p_i =\frac{\nor{X_i}\nor{Y_i}}{S}$, where $S=\sum_{i=1}^n\nor{X_i}\nor{Y_i}$. Form $B_{X}$ and $B_{Y}$ by taking $\ell$ iids samples (column indices), using $p_i$. In the streaming model, since $S$ is not known, we use $\ell$ independent reservoir samples.  Hence the space needed is $O(\ell(m_x+m_y))$, the running time is $O(\ell(m_x+m_y)n)$. 

\textbf{Random Projection }\cite{Sarlos2006}. $B_{X},B_Y$ are of the form $X\Pi$ and $Y \Pi$, where $\Pi \in \mathbb{R}^{n \times \ell}$
, and $\Pi_{ij}\in \{-1/\sqrt{\ell},1/\sqrt{\ell}\}$, uniformly. This is easily implemented in the streaming model and requires $O(\ell(m_x+m_y))$ space and $O(\ell(mx+m_y)n)$ time.

\textbf{Hashing } \cite{Clarkson2013}. Let $h: [n]\to [\ell]$, and $s: [n] \to \{-1,1\}$ be perfect hash functions. We initialize $B_{X},B_{Y}$ to all zeros matrices.
When processing columns of $X$ and $Y$ we update columns of $B_X$ and $B_{Y}$ as follows: $B_{X,h(i)}\gets B_{X,h(i)} + s(i)X_i, B_{Y,h(i)}\gets  B_{Y,h(i)} + s(i)Y_i $. Hashing requires $O(\ell(m_x+m_y))$ space and $O(n(m_x+m_y))$ time.

\textbf{FD-AMM } \cite{YeLZ16}. Let $Z=[X;Y] \in \mathbb{R}^{(m_x+my)\times n}$, let $D_{Z}$ be the output of frequent directions (Algoritm \ref{ALG:fd}). We partition $D_Z= [B_X;B_Y]$, and use $B_{X}$ and $B_Y$ in AMM.
This requires $O(\ell(m_x+m_y))$ space and $O(n(m_x+m_y)\ell)$ time.

\section{Experiments}\label{sec:experiments}
%\subsection{Synthetic Data }
%\paragraph{General AMM.}
\vskip -0.1 in
\paragraph{AMM of Low Rank Matrices.} We consider $X \in \mathbb{R}^{m_x \times n}$ and $Y \in \mathbb{R}^{m_y\times n}$, generated using a non-noisy low rank model \cite{FD} as follows: $X=V_xS_xU^{\top}_{x}$, where $U_x\in \mathbb{R}^{n\times k_x}$, $(U_{x})_{i,j}\sim \mathcal{N}(0,1)$, $S_x\in \mathbb{R}^{k_x\times k_x}$ is a diagonal matrix with $(S_{x})_{jj}=1-(j-1)/k_x$, and $V_x \in \mathbb{R}^{m_{x}\times k_x}$ is such that $V^{\top}_{x}V_{x}=I_{k_x}$. Similarly we generate $Y =V_yS_yU^{\top}_{y}, U_{y}\in \mathbb{R}^{n\times k_{y}},S_{y}\in \mathbb{R}^{k_y\times k_{y}}, V_y \in \mathbb{R}^{m_y\times k_y}$. Hence $X$ and $Y$ are at most rank $k_x$, and $k_y$ respectively. We consider $n=10000$, $m_x=1000$, $m_y=2000$, and three regimes: both matrices have a large rank $(k_x=400,k_y=400)$, one matrix has a smaller rank then the other $(k_x=400,k_y=40)$, and  both matrices have a small rank $(k_x=40,k_y=40)$. We compare the performance of co-occuring directions to baselines given in Section \ref{sec:prev} in those three regimes. For randomized baselines we run each experiments $50$ times and report mean and standard deviations of performances. Experiments were conducted  on a single core Intel Xeon CPU E5-2667, 3.30GHz, with 265 GB of RAM and 25.6 MB of cache.
\begin{figure}[ht!]
%\hspace*{-0.35in}
\centering
\includegraphics[scale=0.3]{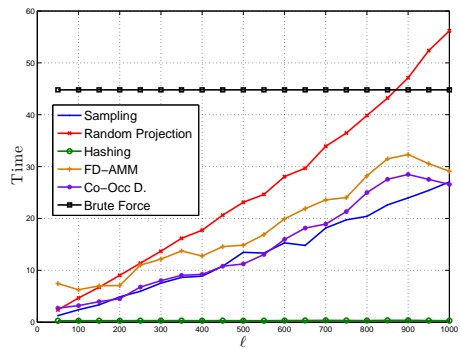}
\caption{Time given in seconds versus sketch length $\ell$.}
\label{fig:timeExp}
\end{figure}

\begin{figure*}[ht!]
\hspace*{0.35in}
   \begin{subfigure}[t]{0.4\textwidth}        
   \includegraphics[scale=0.3]{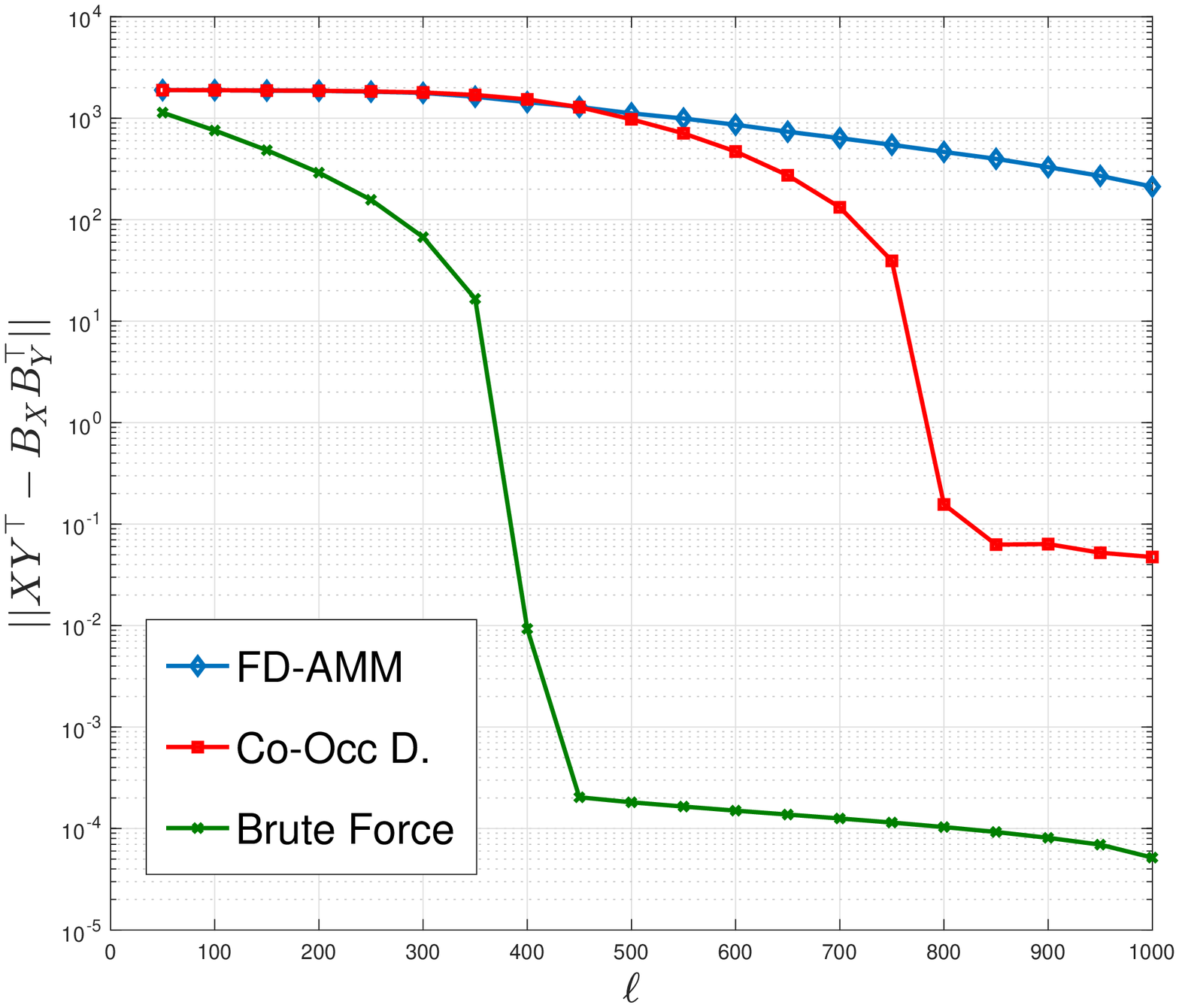}
  %  \hspace*{2.8in}
 \caption{no noise $(k_x=400,k_y=400)$,\\ error in $\log$ scale.}
    \end{subfigure}
 ~
   \hspace*{0.35in}
    \begin{subfigure}[t]{0.45\textwidth} 
    \includegraphics[scale=0.32]{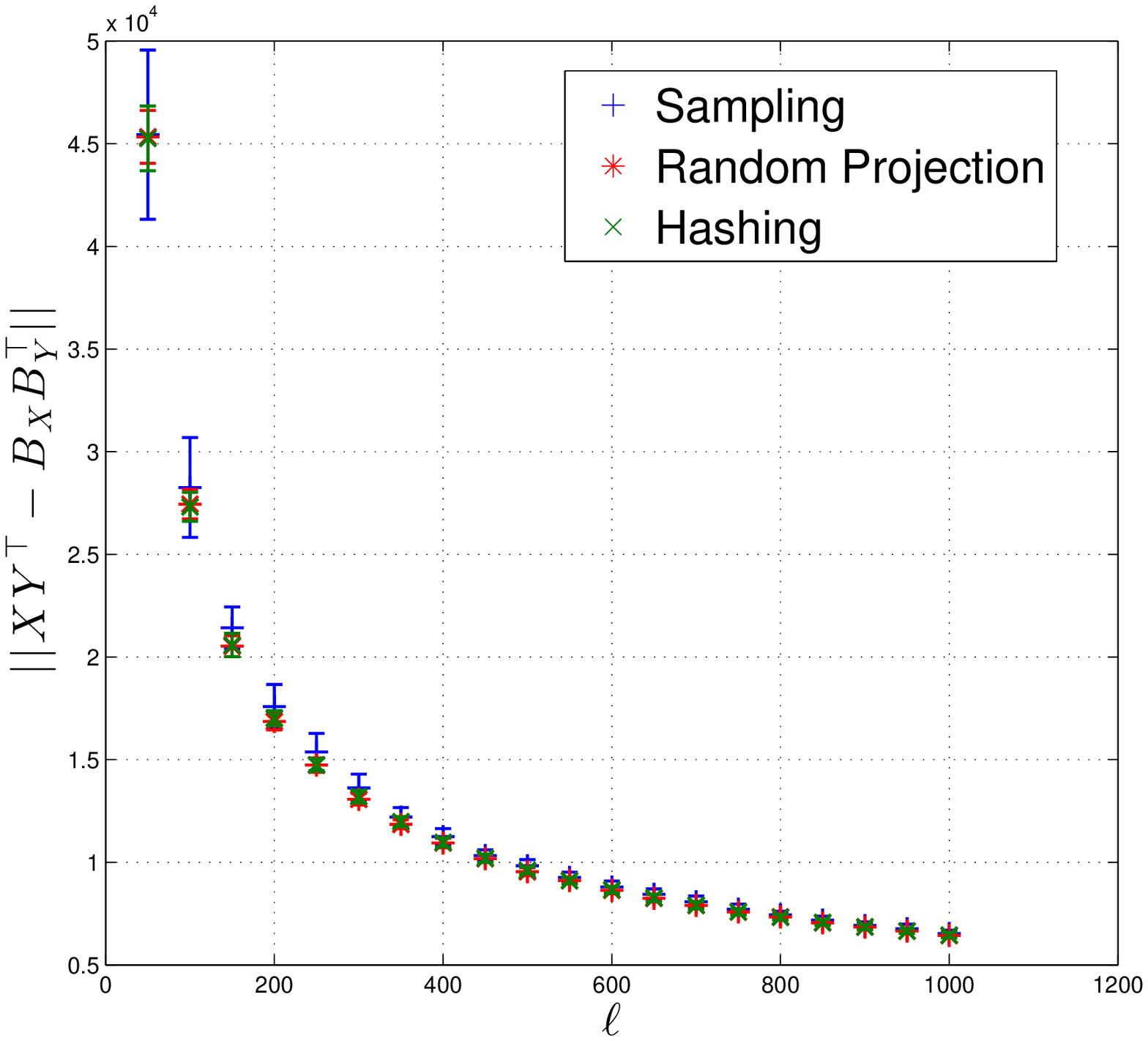}
    %\hspace*{-0.6in}
     %  \centering
         \caption{no noise $(k_x=400,k_y=400)$ \\error in linear scale. }          
             \end{subfigure}
~
\hspace*{0.35in}
    \begin{subfigure}[t]{0.4\textwidth} 
    \includegraphics[scale=0.3]{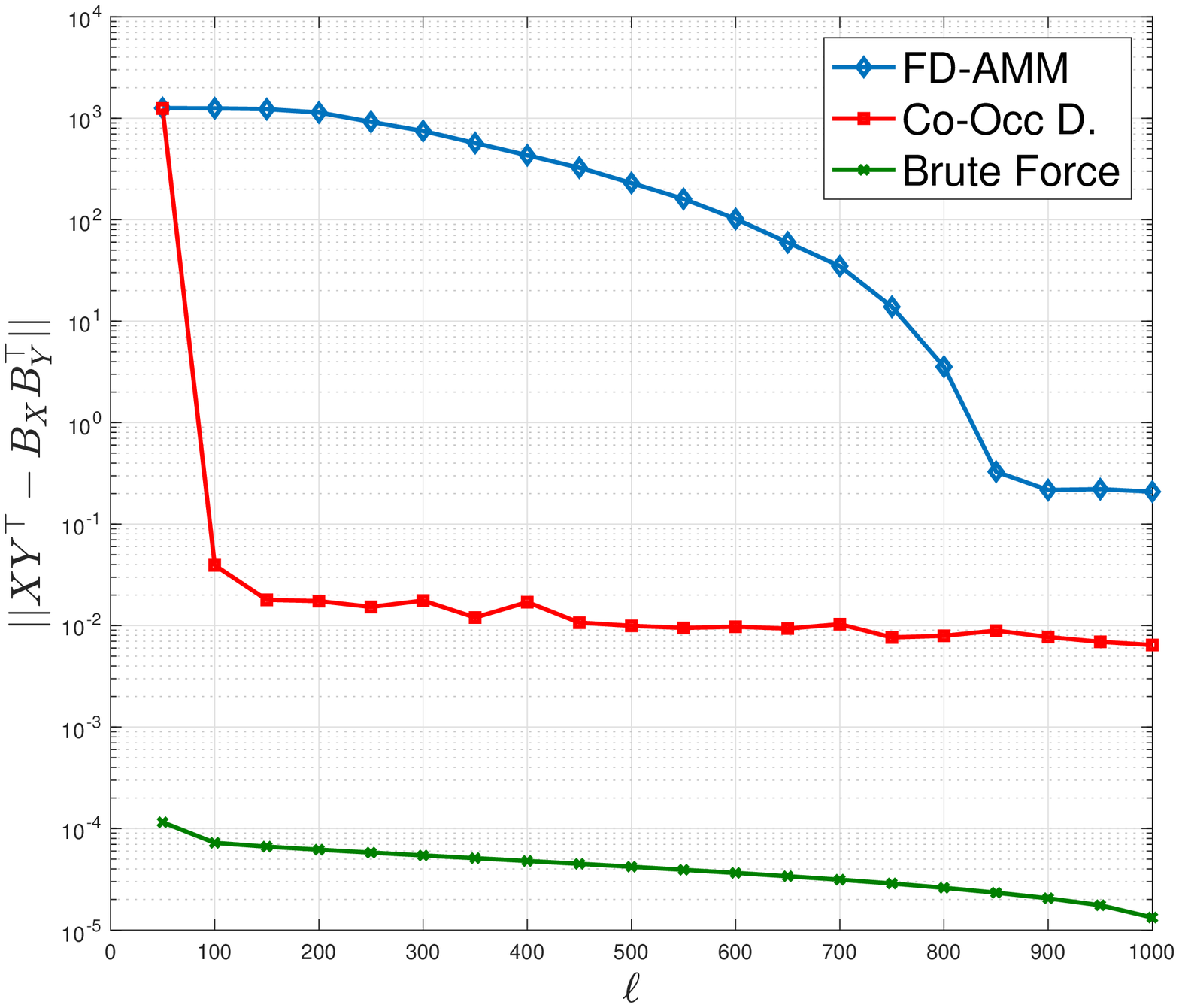}
    %\hspace*{-0.6in}
     %  \centering
         \caption{no noise $(k_x=400,k_y=40)$\\error in $\log$ scale. }
    \end{subfigure} 
         ~       
    \hspace*{0.35in}
    \begin{subfigure}[t]{0.45\textwidth} 
    \includegraphics[scale=0.32]{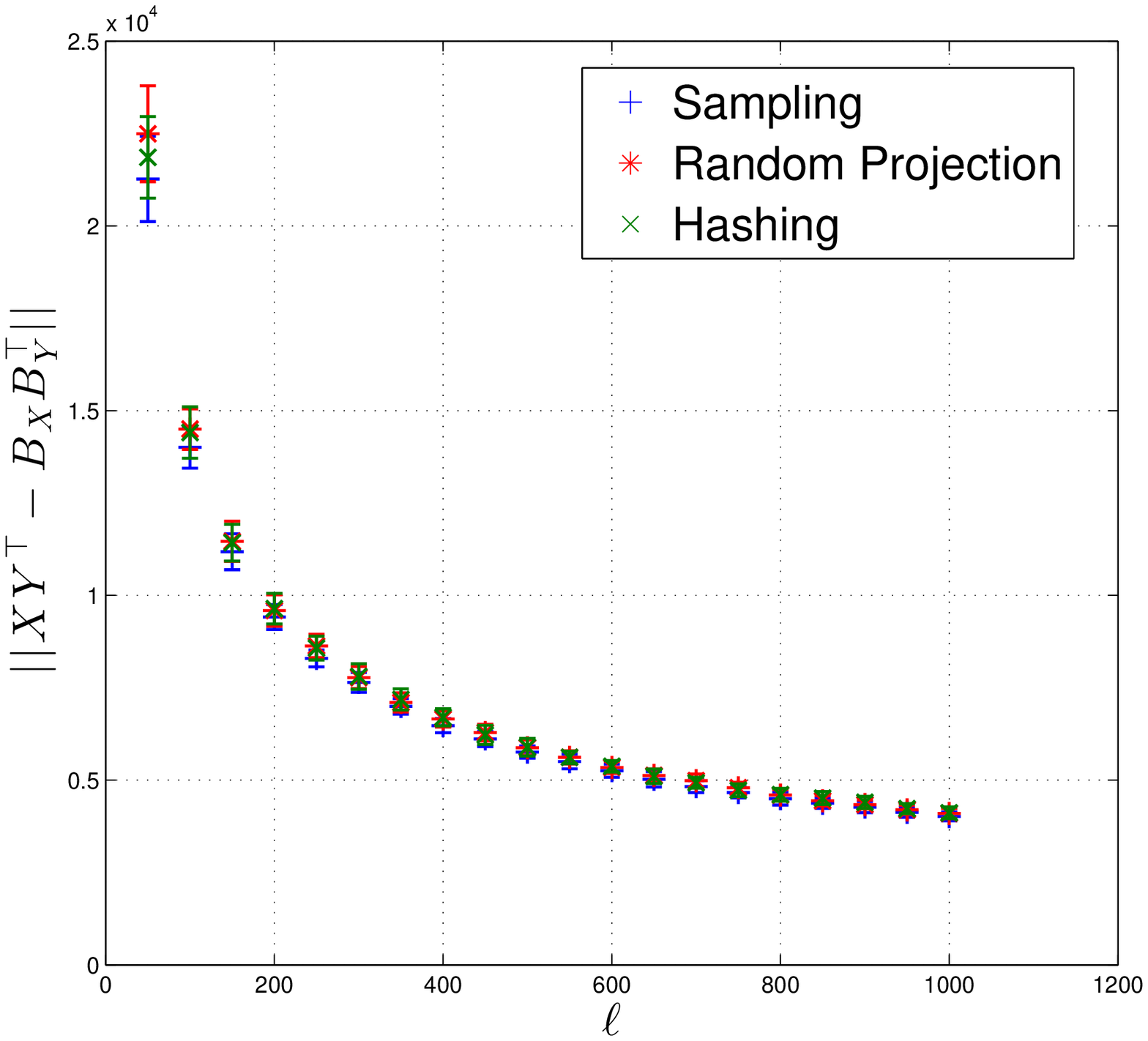}
    %\hspace*{-0.6in}
     %  \centering
         \caption{no noise $(k_x=400,k_y=40)$\\ error in linear scale. }
    \end{subfigure}     
    ~  
    \hspace*{0.35in}
    \begin{subfigure}[t]{0.4\textwidth} 
    \includegraphics[scale=0.30]{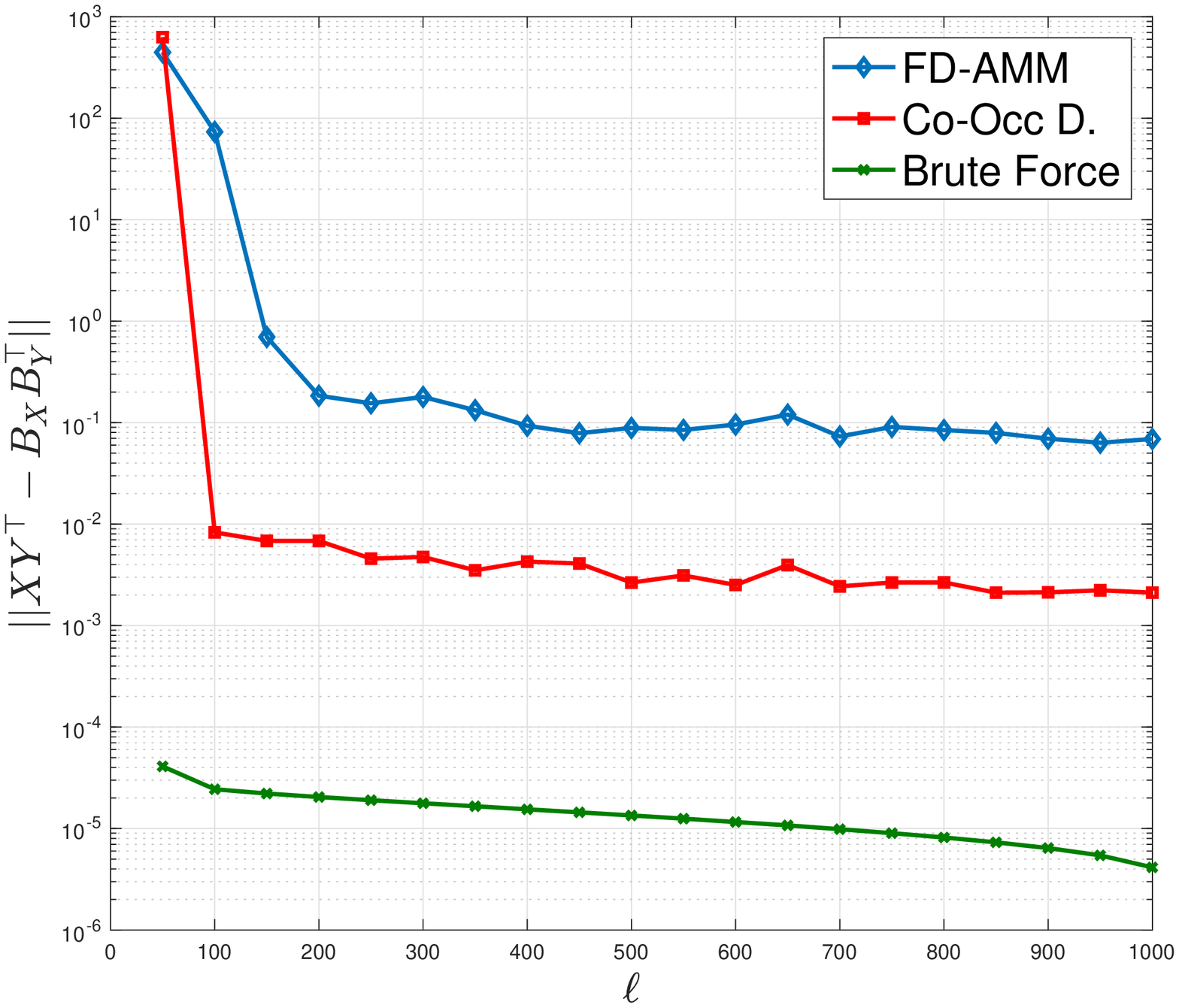}
    %\hspace*{-0.6in}
     %  \centering
         \caption{no noise $(k_x=40,k_y=40)$\\ error in $\log$ scale. }
    \end{subfigure}     
           ~       
       \hspace*{0.35in}
    \begin{subfigure}[t]{0.45\textwidth} 
    \includegraphics[scale=0.33]{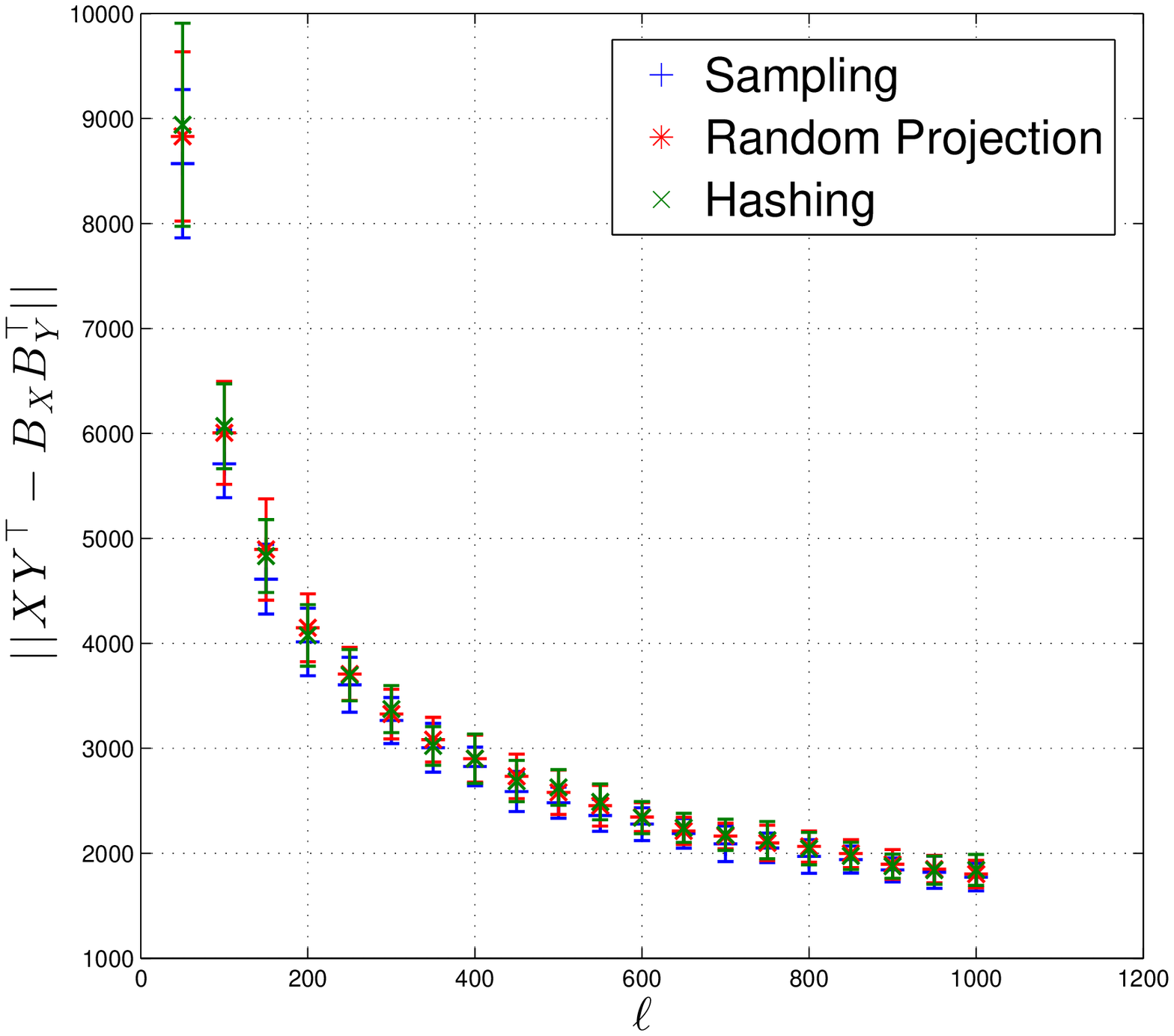}
    %\hspace*{-0.6in}
     %  \centering
         \caption{no noise $(k_x=40,k_y=40)$\\ error in linear scale. }
    \end{subfigure}     
~
  \caption{(a),(c),(e)Error of co-occuring directions versus the deterministic baseline FD-AMM, for clarity the error is given in log scale. (b)(d)(f) Error of co-occuring directions versus randomized baselines (sampling, random projection and hashing), for clarity the error is given in linear scale.}
  \label{fig:csketch}   
  \end{figure*} 
  \vskip -0.2 in 
 We see in Figure \ref{fig:timeExp}, that hashing timing is, as expected,  independent from the sketch length. Random projection requires the most amount of  time. Co-occuring directions timing is on par with sampling and slightly better than FD-AMM. From Figure \ref{fig:csketch} \footnote{Better seen in color.} we see that the deterministic baselines (a,c,e)  consistently outperform the randomized baselines (b,d,f) in all three regimes. As discussed previously randomized methods error bound are of the order of $O(1/\sqrt{\ell})$, while both co-occuring directions and FD-AMM have an error bound order $O(1/\ell)$. Note that the brute force error becomes zero (up to machine precision) when $\ell$ exceeds $\min(rank(X),rank(Y))$. When comparing co-occuring direction to FD-AMM we see a clear phase transition for co-occuring direction as $\ell$ exceeds $O(\min(rank(X),rank(Y)))$. For FD-AMM the phase transition happens when $\ell$ exceeds $O(rank(X)+rank(Y))$. The phase transition happens earlier for co-occuring directions and hence co-occuring directions outperforms FD-AMM for a smaller sketch size. This is in line with our discussion in Section \ref{sec:discussion}.
For instance plot (c) illustrates this effect, $k_x=400,k_y=40$, as $\ell$ exceeds $50$, the error of co-occuring directions  sharply decreases , while FD-AMM error is still high. The latter starts a steep decreasing tendency when $\ell$ exceeds $400$. We give plots for the low rank approximation as given in Theorem \ref{theo:LR} for $k=\min(k_x,k_y)$ in  the appendix, we see a similar trend in the approximation error.

\begin{figure}[h]
%\hspace*{0.35in}
 %  \begin{subfigure}[t]{0.3\textwidth}  
 \centering      
   \includegraphics[scale=0.28]{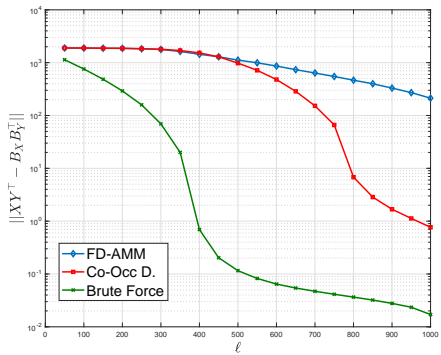}
  %  \hspace*{2.8in}
 \caption{Noisy $(k_x=400,k_y=400)$. $\log$ scale.}
   %\end{subfigure}
   \label{fig:fig3}
 \end{figure}

\begin{figure}[h]
%\hspace*{0.35in}
 %  \begin{subfigure}[t]{0.3\textwidth}  
  \centering         
   \includegraphics[scale=0.28]{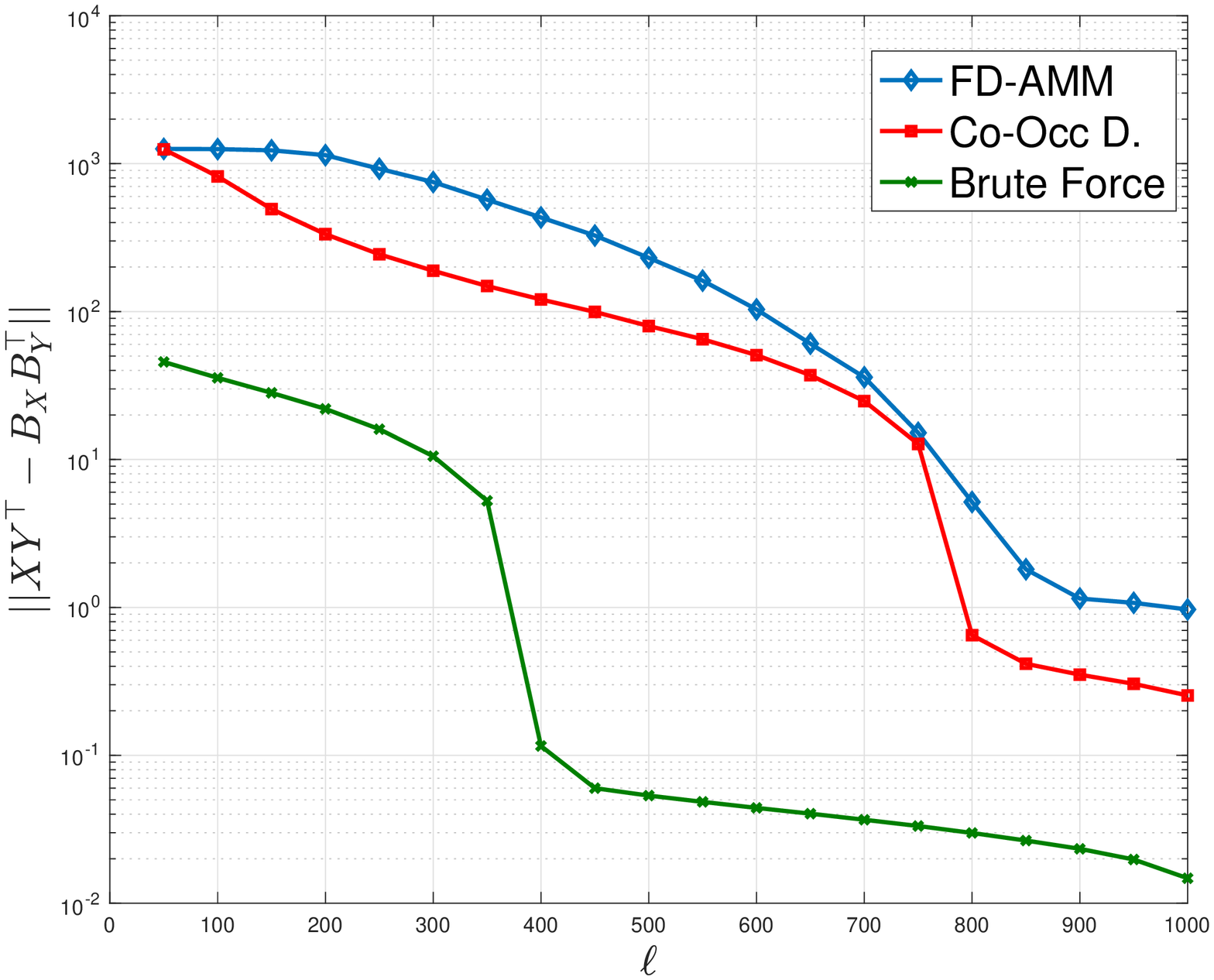}
  %  \hspace*{2.8in}
 \caption{Noisy$(k_x=400,k_y=40)$. Error in $\log$ scale.}
    \label{fig:fig4}
   %\end{subfigure}
 \end{figure}

\begin{figure}[h]
%\hspace*{0.35in}
 %  \begin{subfigure}[t]{0.3\textwidth}    
  \centering       
   \includegraphics[scale=0.28]{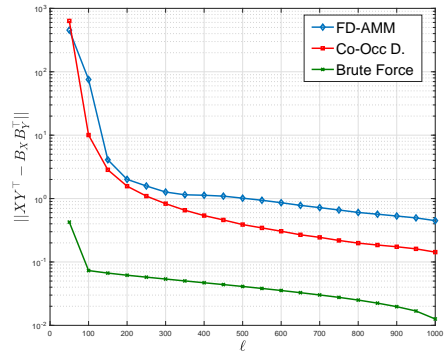}
  %  \hspace*{2.8in}
 \caption{Noisy $(k_x=40,k_y=40)$. Error in $\log$ scale.}
   %\end{subfigure}
      \label{fig:fig5}
 \end{figure}

\textbf{AMM of Noisy Low Rank Matrices (Robustness).} We consider the same model as before but we add a gaussian noise to the low rank matrices, i.e $X =V_{x}S_{x}U^{\top}_{x}+N_x/\zeta_x $, where $\zeta_x>0$, and $N_{x}\in \mathbb{R}^{m_x\times n}$, $(N_x)_{i,j}\sim \mathcal{N}(0,1)$. Similarly for $Y=V_{y}S_{y}U^{\top}_{y}+ N_y/\zeta_y$. In this scenario $X$ and $Y$  have still decaying singular values but with non zeros tails.  We consider $\zeta_x=1000$, and $\zeta_y=100$. We compare here deterministic baselines in Figures \ref{fig:fig3},\ref{fig:fig4}, and \ref{fig:fig5}, in the three scenarios we see that co-occuring directions still outperforms FD-AMM, but the gap between the two approaches becomes smaller in the low rank regimes (Figures \ref{fig:fig4}, and \ref{fig:fig5}), this hints to a weakness in the shrinking of singular values in both algorithms getting affected by the noise (Step 17 in Alg. \ref{ALG:CorrSketch}). We give plots for the low rank approximation in  the appendix.
\vskip -0.1 in
\vskip -0.2 in

\vskip -0.1 in
 
\begin{figure}[h]
%\hspace*{0.35in}
 %  \begin{subfigure}[t]{0.3\textwidth}  
  \centering         
   \includegraphics[scale=0.28]{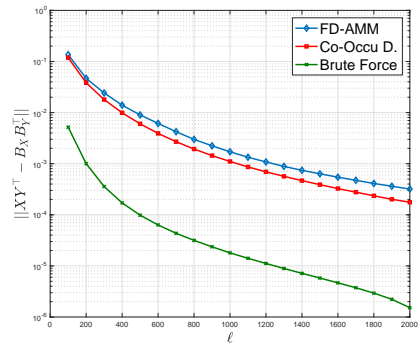}
  %  \hspace*{2.8in}
 \caption{AMM error on MS-COCO.}
    \label{fig:fig6}
   %\end{subfigure}
 \end{figure}

 \textbf{Multimodal Data Experiments.} In this section we study the empirical performance of co-occuring directions in approximating correlation between images and captions. We consider Microsoft COCO \cite{coco} dataset. For visual features we use the residual CNN Resnet101, \cite{he15deepresidual}. The last layer of Resnet results in  a feature vector  of dimension $m_x= 2048$.  For text we use the Hierarchical Kernel Sentence Embedding HSKE of \cite{assymetric} that results in a feature vector of dimension $m_y=3000$. 
The training set size is $n=113287$. We see in Fig. \ref{fig:fig6} that co-occuring directions outperforms FD-AMM in this case as well (timing experiment is given in the appendix).
\vskip -0.1 in

 %%   \hspace*{0.35in}
%    \begin{subfigure}[t]{0.3\textwidth} 
%    \includegraphics[scale=0.35]{./epfig/error_400_40_1000_100.eps}
%    %\hspace*{-0.6in}
%     %  \centering
%         \caption{$(k_x=400,k_y=400)$\\ error in linear scale. }          
%             \end{subfigure}
%~
%%\hspace*{0.35in}
%    \begin{subfigure}[t]{0.3\textwidth} 
%    \includegraphics[scale=0.35]{./epfig/error_40_40_1000_100.eps}
%    %\hspace*{-0.6in}
%     %  \centering
%         \caption{$(k_x=400,k_y=40)$\\error in $\log$ scale. }
%    \end{subfigure} 
%
%~
%  \caption{}
%  \label{fig:csketch}   
%  \end{figure} 
%\subsection{AMM and Multimodal Retrieval }

\section{Conclusion}
In this paper we introduced  a deterministic sketching algorithm for AMM that we termed co-occuring directions . We showed its error bounds (in spectral norm) for AMM  and the low rank approximation of a product. We showed empirically that co-occuring directions outperforms deterministic and randomized baselines in the streaming model. Indeed co-occuring direction has the best error/space tradeoff among known baselines with errors given in spectral norm in the streaming model.
We are left with two open questions. First, whether guarantees of Theorem \ref{theo:SketchCorr} can be improved akin to the improved guarantees for \emph{frequent directions} given \cite{FD}. This would give an explicit link of the sketch length $\ell$, to the low rank structure of the matrix product $XY^{\top}$, and/or the low rank structure of the individual matrices. Second, whether robustness of co-occuring directions can be improved using  robust shrinkage operators as in \cite{Ghashami2014}.

\bibliographystyle{alpha}
 \bibliography{simplex}

\onecolumn
\appendix
\section{Low Rank product Approximation}\label{app:LR}
%\onecolumn
\begin{figure}[ht!]
%\hspace*{0.35in}
   \begin{subfigure}[t]{0.4\textwidth} 
   \centering       
   \includegraphics[scale=0.3]{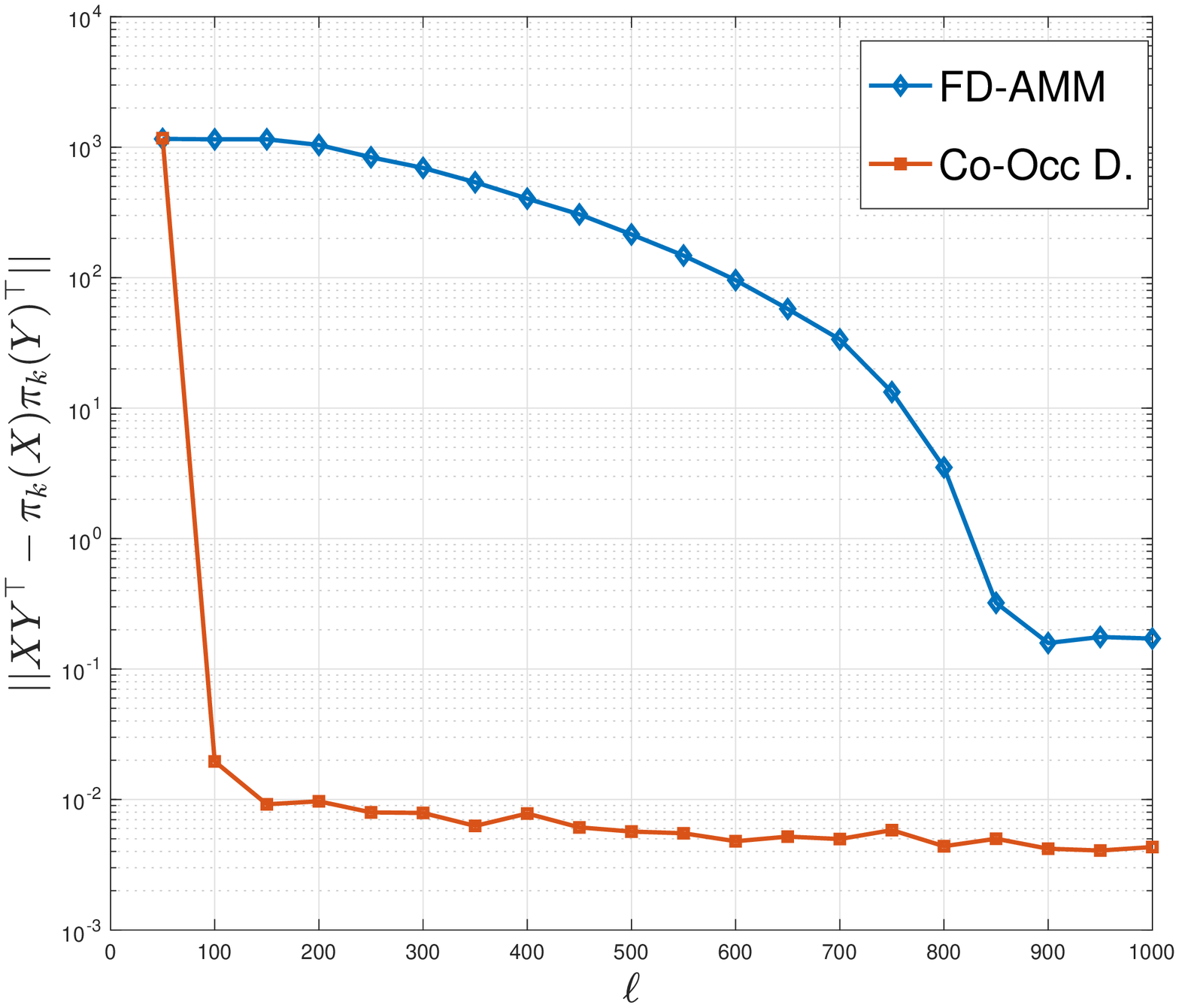}
  %  \hspace*{2.8in}
 \caption{$(kx=400,k_y=40)$\\ error in $\log$ scale.}
    \end{subfigure}
 ~
   %\hspace*{0.35in}
    \begin{subfigure}[t]{0.45\textwidth} 
    \includegraphics[scale=0.3]{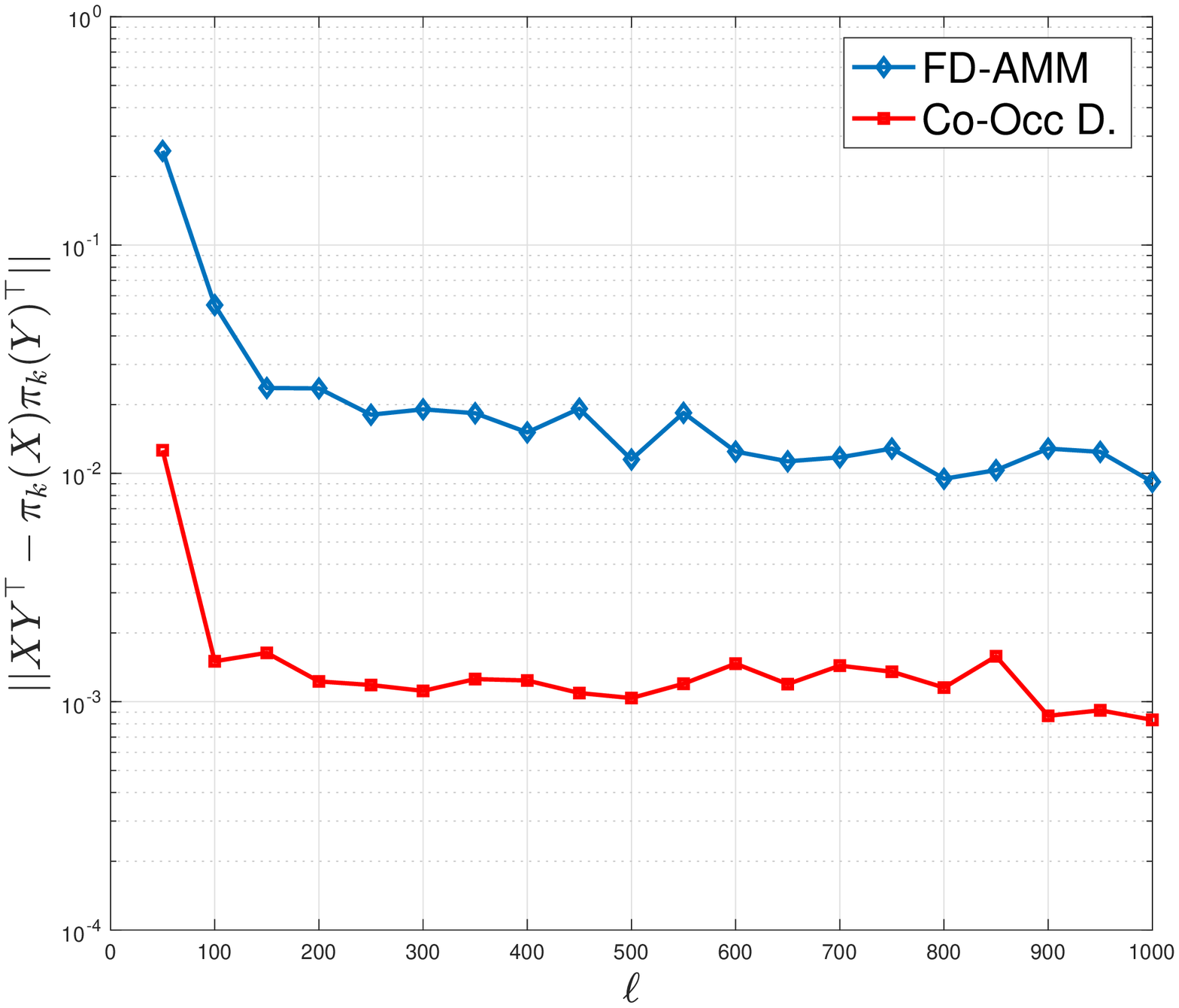}
    %\hspace*{-0.6in}
      \centering
         \caption{$(k_x=40,k_y=40)$\\ error in $\log$ scale. }          
             \end{subfigure}

  \caption{No noise : Low rank approximation of matrix product, after projection on left and right singular vectors of $B_{X}B_{Y}^{\top}$ for $k=\min(k_x,k_y)=40$.}
  \label{fig:csketchLR}   
  \end{figure}

  \begin{figure}[ht!]

   \begin{subfigure}[t]{0.4\textwidth}        
\centering
   \includegraphics[scale=0.32]{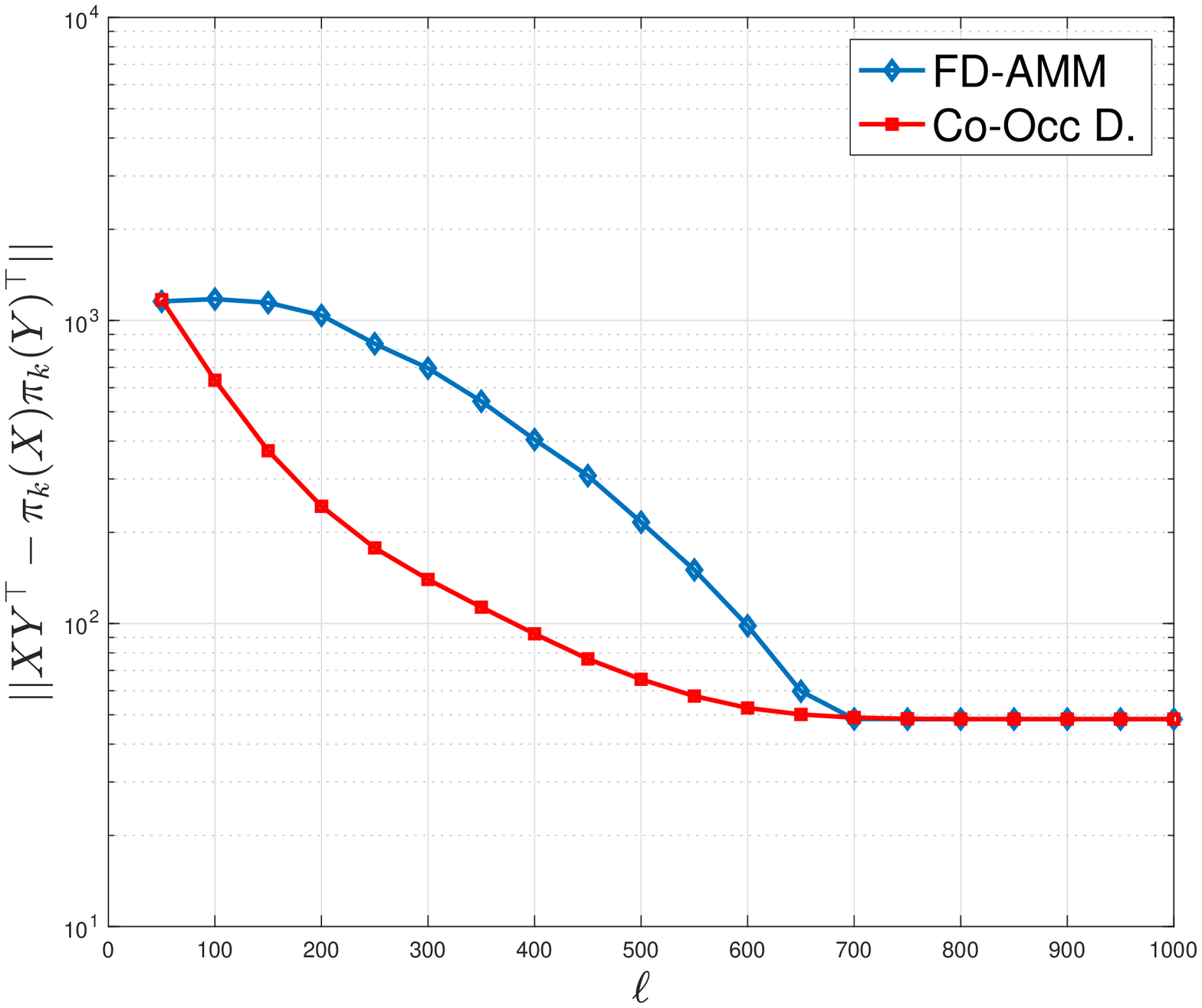}
  %  \hspace*{2.8in}
 \caption{$(kx=400,k_y=40)$\\ error in $\log$ scale.}
    \end{subfigure}
 ~
 %  \hspace*{0.45in}
    \begin{subfigure}[t]{0.45\textwidth} 
    \centering

    \includegraphics[scale=0.32]{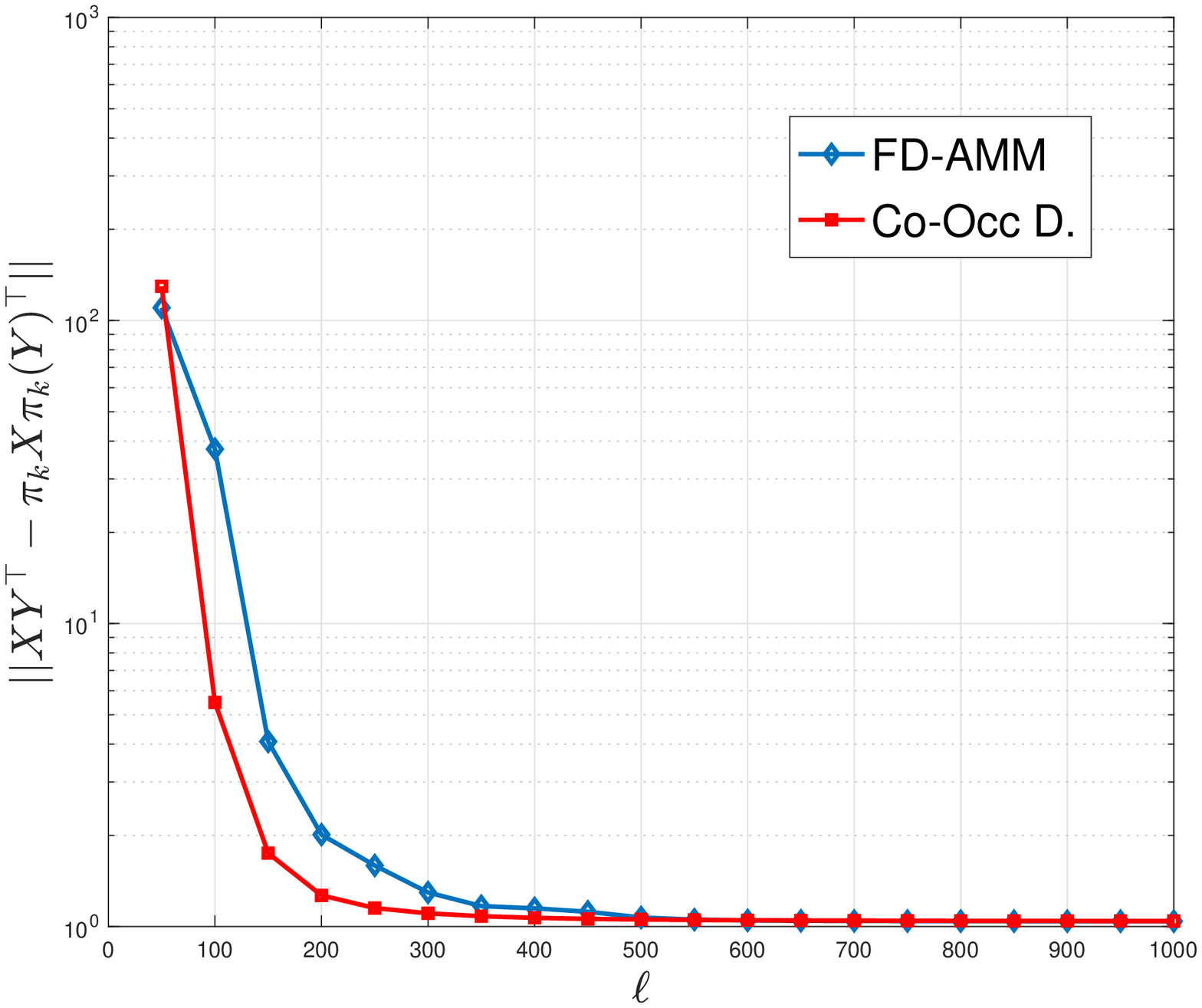}
    %\hspace*{-0.6in}
     %  \centering
         \caption{$(k_x=40,k_y=40)$\\ error in $\log$ scale. }          
             \end{subfigure}

  \caption{Noisy : Low rank approximation of matrix product, after projection on left and right singular vectors of $B_{X}B_{Y}^{\top}$ for $k=min(k_x,k_y)=40$.}
  \label{fig:csketchNOISE}   
  \end{figure}

  \section{MS-Coco Timing Experiments}
  \begin{figure}[ht!]
%\hspace*{0.35in}
 %  \begin{subfigure}[t]{0.3\textwidth}  
  \centering         
   \includegraphics[scale=0.34]{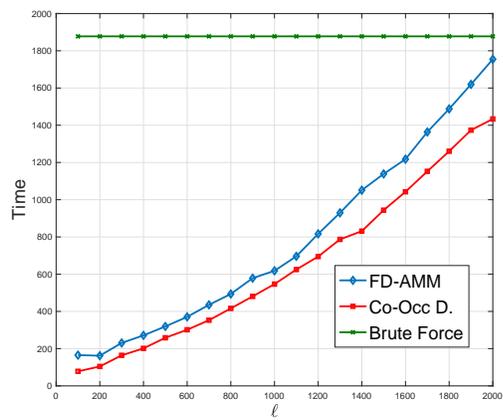}
  %  \hspace*{2.8in}
 \caption{Timing of sketching on MS-COCO.}
    \label{fig:timecoco}
   %\end{subfigure}
 \end{figure}

\end{document}